\pdfoutput=1

\documentclass[11pt]{article}

\usepackage[final]{acl}

\usepackage{times}
\usepackage{latexsym}

\usepackage[T1]{fontenc}

\usepackage[utf8]{inputenc}

\usepackage{microtype}

\usepackage{inconsolata}

\usepackage{graphicx}
\usepackage{amsmath}

\usepackage{adjustbox}
\usepackage{booktabs}
\usepackage{algorithm}
\usepackage{algpseudocode}
\usepackage{tabularx}
\usepackage{makecell}
\usepackage{amssymb}
\usepackage{pifont}
\usepackage{colortbl} 
\usepackage{xcolor}   
\newcommand{\cmark}{\ding{51}}
\newcommand{\xmark}{\ding{55}}

\usepackage{multirow}

\title{DynamicER: Resolving Emerging Mentions to Dynamic Entities for RAG}

\author{Jinyoung Kim \qquad Dayoon Ko \qquad Gunhee Kim \\
        \\[-4mm]
        Seoul National University \\
        \small{\texttt{jiny1623@snu.ac.kr}} \quad \small{\texttt{dayoon.ko@vision.snu.ac.kr}} \quad \small{\texttt{gunhee.kim@snu.ac.kr}} \\
        \small{\href{https://github.com/jiny1623/DynamicER}{\texttt{https://github.com/jiny1623/DynamicER}}}
        }

\definecolor{check}{rgb}{0.333, 0.755, 0.545}

\begin{document}
\maketitle
\begin{abstract}

In the rapidly evolving landscape of language, resolving new linguistic expressions in continuously updating knowledge bases remains a formidable challenge. 
This challenge becomes critical in retrieval-augmented generation (RAG) with knowledge bases, as emerging expressions hinder the retrieval of relevant documents, leading to generator hallucinations. 
To address this issue, we introduce a novel task aimed at resolving emerging mentions to dynamic entities and present \textproc{DynamicER} benchmark. Our benchmark includes dynamic entity mention resolution and entity-centric knowledge-intensive QA task, evaluating entity linking and RAG model's adaptability to new expressions, respectively. 
We discovered that current entity linking models struggle to link these new expressions to entities. Therefore, we propose a temporal segmented clustering method with continual adaptation, effectively managing the temporal dynamics of evolving entities and emerging mentions. Extensive experiments demonstrate that our method outperforms existing baselines, enhancing RAG model performance on QA task with resolved mentions. 

\end{abstract}

\section{Introduction}

\begin{figure}[t!]
    \includegraphics[width=0.48\textwidth]{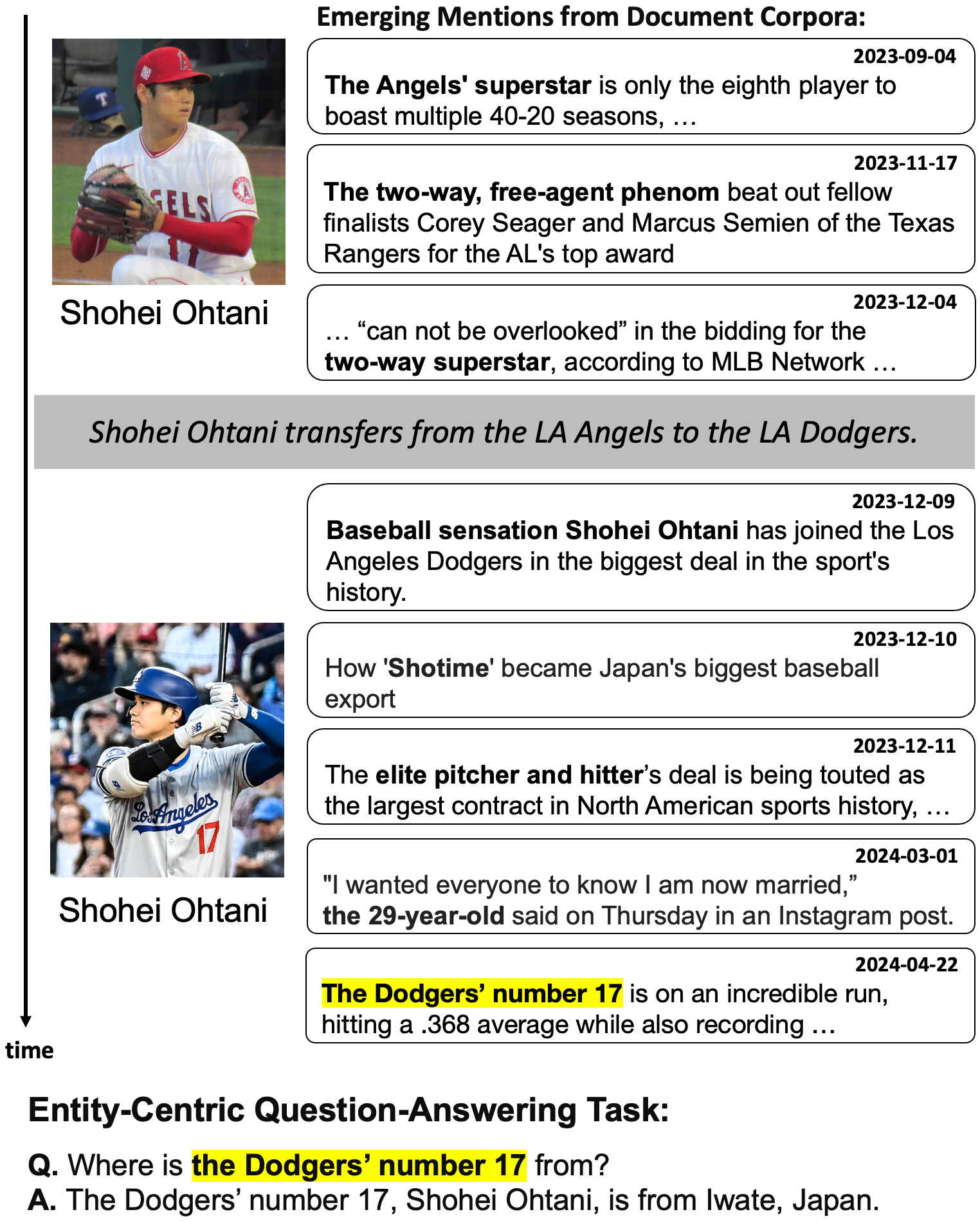}
    \caption{Motivation of our \textproc{DynamicER} benchmark. New mentions referring to the same entity are constantly created over time: as Shohei Ohtani transfers from the LA Angels to the LA Dodgers, he is referred to by new mentions such as ‘The Dodgers' number 17.’ We contribute a dynamic entity resolution dataset, along with two benchmark tests: traditional entity linking and entity-centric question-answering in the RAG context.}
    \label{fig:figure1}
\end{figure}

In the real world, large amounts of textual information are constantly being generated at an incredible rate. The dynamic nature of human language, characterized by the continuous emergence of new expressions, presents multiple significant challenges \citep{hirschberg2015advances}. The way we refer to named entities changes over time, influenced by the shifts that include the use of metaphors, adoption of slang, creation of euphemisms, and other linguistic evolutions \citep{li2020survey}.
For example, consider the named entity ``Elon Musk''. Over time, various mentions are used to refer to him, such as ``the Tesla CEO'', or ``the tech billionaire''. Emerging slang or metaphoric expressions like ``real-life Iron Man'' or ``Mars man'' also appear. As a dynamic entity, his attributes change over time as he was initially known as "PayPal co-founder'', later as ``Hyperloop visionary'', and more recently as ``Twitter owner''. 
All of these expressions refer to the same entity at different time points, yet a system must be able to recognize them despite the dynamic linguistic landscape. Therefore, it is crucial for a system to resolve these new expressions, accurately linking them to evolving entities in a continuously updating knowledge base (KB). 

Recently, KBs have been critically utilized in the retrieval-augmented generation (RAG) framework \citep{lewis2020retrieval, guu2020retrieval, izacard2023atlas, asai2023self}, where a retriever fetches relevant documents from KBs, allowing a generator like large language models (LLMs) to provide accurate answers. 
As LLMs become capable of learning with prompt, studies have utilized RAG for knowledge-intensive open-domain tasks. 
Following this trend, efforts \citep{liska2022streamingqa,dhingra2022time,neelam-etal-2022-sygma} have been made to manage dynamically evolving knowledge with RAG. In particular, some studies \citep{kasai2024realtime, ko2024growover} have shown that in time-sensitive knowledge-intensive benchmarks, LLMs inevitably produce outdated answers if retrieval fails.
One significant reason for retrieval failure is the emergence of new mentions, which prevents the retriever from functioning properly \citep{sciavolino2021simple,mallen2023not}.
However, to the best of our knowledge, no approach has tackled the challenge of resolving new expressions of named entities evolving over time for the RAG framework. 

To overcome this limitation, we call for a dynamic entity resolution task that links emerging mentions to dynamic entities. We contribute the \textproc{DynamicER} (\textbf{Dynamic} \textbf{E}ntity \textbf{R}esolution for Emerging Mentions) benchmark, designed to study the resolution of continuously evolving entities and their newly appearing mentions.
As illustrated in Figure~\ref{fig:figure1}, \textproc{DynamicER} annotates emerging mentions found in social media documents, linking these mentions to corresponding named entities within a KB.
\textproc{DynamicER} is structured as a sequence of time segments to evaluate models resolving mentions that are not recognized in earlier time steps.
We also introduce a dynamic entity-centric question-answering (QA) task where named entities in question are substituted into emerging mentions over time. 
This QA task examines the impact of emerging mentions on the retriever's accuracy and aims to evaluate the end-to-end performance of the RAG methods. 
Table~\ref{tab:comparison} presents a comparison of \textproc{DynamicER} with other benchmarks.

When a new mention first appears, it may be challenging to resolve it using only a single document since it is likely to have high lexical variation and insufficient context.  
Although prior works have used mentions in multiple documents collectively \citep{ganea-hofmann-2017-deep, le-titov-2018-improving, angell-etal-2021-clustering, agarwal-etal-2022-entity}, 
there is a potential risk when jointly handling mentions from different time steps. The definition or attributes of entities may evolve across time steps, and neighbor mentions referenced together also change accordingly.
To address this issue, we propose a method that continuously clusters entities and a set of mentions at each time step and updates the evolving entity cluster representations.

Our contributions are summarized as follows:
\begin{enumerate}
    \item We introduce \textproc{DynamicER} as the first benchmark for linking emerging mentions to dynamic entities, measuring the capability of RAG models to resolve and adapt to newly emerging expressions.
    \item We propose a temporal segmented clustering with continual adaptation, which considers temporal dynamics to distinguish between entities and mentions more effectively, especially when new mentions emerge.
    \item We empirically show that our method outperforms other entity linking methods and that resolving new mentions is beneficial for RAG performance in QA tasks.
\end{enumerate}

\section{Related Work}

\begin{table*}[t!]
\centering
\begin{adjustbox}{width=\linewidth}
{\small
\begin{tabular}{lccccc}
\toprule
& \textproc{MedMentions} & \textproc{Zero-shot EL} & \textproc{Reddit EL} & \textproc{TempEL} & \textproc{DynamicER} \\
& \citeyearpar{mohan2018medmentions} & \citeyearpar{logeswaran2019zero} & \citeyearpar{botzer2021reddit} & \citeyearpar{zaporojets2022tempel} & (Ours)
\\\midrule
    Source 
        & PubMed & Wikias & Social media & Wikipedia & Social media \\\midrule
    Domain 
        & Biomedical & Fictional universes & General & General & Sports \\\midrule
    Size
        & 
            \begin{tabular}{c}350K \\ (4K Docs)\\\end{tabular} 
        &    
            \begin{tabular}{c}70K \\ (16 Wikias)\\\end{tabular}
        & 
            \begin{tabular}{c}17K \\ (619 Posts)\\\end{tabular}
        & 
            240K
        & 
            \begin{tabular}{c}70K \\ (20K Docs)\\\end{tabular}
        \\\midrule
    Temporal dynamics
        & \xmark & \xmark & \xmark & {\color{check}\cmark} & {\color{check}\cmark} \\\midrule
    Mention variations
        & {\color{check}\cmark} & {\color{check}\cmark} & {\color{check}\cmark} & \xmark & {\color{check}\cmark} \\\midrule
    Tasks
        & Entity linking & Entity linking & Entity linking & 
        \begin{tabular}{c}Continuous \\entity linking \end{tabular}& 
        \begin{tabular}{c}Dynamic entity \\ mention resolution \\ \& Entity-centric QA \end{tabular}\\\bottomrule
    \end{tabular}}
    \end{adjustbox} 
    \caption{
    Comparison of our \textproc{DynamicER} with existing entity linking benchmarks. The source indicates from which the dataset is collected. The domain shows the type of data content. The size displays the number of mentions along with the number of documents, Wikias, or posts. The temporal dynamics represent whether the dataset evolves or changes over time. The mention variations show whether the dataset is annotated with alias lists of mentions.
    }
    \label{tab:comparison}
\end{table*}

\textbf{Entity Linking}. 
Entity linking \citep{hoffart2011robust, guo2018robust} aims to match an entity mention to a unique named entity in a KB such as Wikipedia pages. 
There has been much research on how to correctly link varied mentions of the same entity.
For instance, \citet{andy-etal-2017-constructing} design an algorithm to identify entities from social media during the 3-4 hour Grammy Awards, constructing an alias list for short-term use.
\citet{botzer2021reddit} collect a Reddit entity linking dataset, demonstrating that models trained on conventional text encounter difficulties with the unique formats and lexical variations prevalent in social media.
In the biomedical domain, \citet{mohan2018medmentions} propose the MedMentions dataset, which compiles a comprehensive biomedical corpus with entity mention annotations. 
However, the static nature of this domain fails to accommodate the time-evolving linguistic evolution.
The zero-shot setting of entity linking \citep{lin-etal-2017-list, logeswaran2019zero} targets linking entities unseen in training time. This task focuses on domain adaptation to resolve new entities, rather than on handling the mention variation of specific entities over time.

For temporal entity linking, the TempEL dataset offers detailed tracking and annotation of changes in existing entities as well as the emergence of new entities across multiple temporal snapshots. Our annotation is similar to TempEL in that we handle temporal development of existing entities. However, our dataset also considers the continuous emergence of new mentions of the same entity, highlighting that the way it is referenced varies across multiple points in time.

\textbf{Coreference Resolution}.
Coreference resolution (CR) \citep{pradhan2012conll, webster2018mind} evaluates a model's ability to match entities with their antecedents. The task is to group all spans that point to the same objects in a context by detecting mentions. \citet{lee2017end} integrates these two processes in an end-to-end manner by considering all spans as potential coreference candidates and learning a conditional probability distribution for clustering.  \citet{joshi2020spanbert} extends BERT by training via masked contiguous random spans and predicting the spans using boundary representation. 

Cross-document CR (CDCR) \citep{cybulska2014using, webster2018mind} resolves coreference across multiple documents. \citet{caciularu2021cdlm} utilizes long-range transformers to encode multiple related documents. \citet{allaway2021sequential} sequentially adds each mention to cluster candidates while incrementally updating the coreference candidate cluster representation. Our dynamic entity linking seems similar to CDCR in that both link the mentions referring to the same entity over documents. However, our task is different since it aims to link varied mentions to the evolving entities of a continuously updating KB.

\textbf{RAG with Dynamic Corpus}.
Since continuously retraining LLMs with up-to-date data is demanding, RAG has been employed to handle temporal adaptability.
For instance, \citet{liska2022streamingqa} and \citet{kasai2024realtime} respectively propose dynamic QA tasks with time-stamped and newly published news articles, demonstrating that retrieving up-to-date documents can improve generation results. Moreover, \citet{dhingra2022time} and \citet{margatina2023dynamic} introduce a cloze query to evaluate the acquisition of temporal knowledge. \citet{neelam-etal-2022-sygma} proposes Knowledge Base QA (KBQA) tasks to evaluate the ability of temporal reasoning. Recently, \citet{ko2024growover} introduces dynamically evolving open-domain QA and dialogue benchmarks along with a novel training-free retrieval-interactive LLM framework. 
While existing works focus on the temporal adaptability of models for retrieval and reasoning, they do not specifically address the dynamic nature of entity expressions, which is crucial for applications requiring precise entity linking over time.

\section{The \textproc{DynamicER} Dataset}

\textproc{DynamicER} consists of two tasks: an entity linking task and an entity-centric QA task. The entity linking task focuses on resolving emerging expressions that appear over time to entities in a KB, while the entity-centric QA task evaluates RAG models in answering entity-specific questions. We construct \textproc{DynamicER} through a pipeline, whose key idea at each stage is to first generate automatically using LLMs, and then thoroughly verify the quality with human review: (1) selecting and filtering textual corpora (\S~\ref{subsec:post-processing}), (2) identifying mentions of target entities (\S~\ref{subsec:mention_id}), (3) annotating the appropriate entity for each mention (\S~\ref{subsec:entity_annotate}), and (4) generating QA pairs using resolved entities (\S~\ref{subsec:entity_centric_qa}). The prompt templates for dataset generation are provided in Appendix~\ref{ex:prompt}.

\subsection{Corpora Collection}\label{subsec:post-processing}

\textbf{Post Selection}. We choose the sports domain to capture the mention variations of famous athletes, teams, and coaches, given its inherently dynamic nature.
This domain is particularly suitable since it features diverse naming conventions, such as nicknames and abbreviations for entities. Moreover, frequent updates and news about events like matches and player transfers contribute to its dynamic nature. The sports domain is also event-driven with clear temporal markers like seasons and tournaments.

We target soccer and baseball for corpora selection.
Specifically, for soccer, we select the top 15 teams according to \textit{Forbes World's Most Valuable Soccer Teams}\footnote{
\url{https://www.forbes.com/lists/soccer-valuations/}} and leagues to which these teams belong. For baseball, we select 30 teams from Major League Baseball. Using the \verb|/tagged| method in Tumblr API, we download all posts tagged with our selected hashtags, focusing on posts from  2023-05-01 to 2024-04-30. The full list of hashtags can be found in Appendix~\ref{ex:tagList}.

\textbf{Initial Filtering}. We filter the posts with fewer than 50 or more than 3000 characters to exclude content that is either too brief to provide sufficient contextual meaning or too lengthy to potentially divert attention with extraneous information.
Additionally, we use the FastText module \citep{joulin2016fasttext, joulin2016bag} to ensure text is written in English, filtering out the posts that are not confidently identified as English.

\subsection{Mention Identification}\label{subsec:mention_id}

We first identify expressions referring to named entities using the GPT-4 turbo \citep{achiam2023gpt}. The prompt we use is as follows: 
`\textit{Please identify all expressions in the given text that explicitly name or 
describe a player, coach, or team. This
includes direct names, nicknames, and any
role-specific references (like positions
or accolades) that refer to a particular
individual or team.}'
We refrain from using typical named entity recognition (NER) models since they struggle to identify long expressions, such as ``\textit{the defending National League champion}''.
We ignore posts for which GPT-4 detects fewer than two expressions, as they lack sufficient contextual information for a resolution dataset.

\subsection{Entity Annotation}\label{subsec:entity_annotate}

Once expressions are mined by GPT-4, we use a substring matcher to highlight each expression in order, followed by human verification. 
We employ a dedicated team of workers to annotate the data. Please refer to Ethics Statement for the details.
We instruct the annotators to adjust the offset of the expression if the highlighting is incorrect and to additionally highlight any missing expressions that refer to players, coaches, or teams.
Next, the annotators link each expression to the corresponding Wikipedia entity. They search Wikipedia for a suitable entity and submit a valid URL of the Wikipedia page to the system.  
If a valid Wikipedia entity cannot be found, or if the context from the post is too ambiguous to resolve the expressions, annotators label it as NOT VALID, which is then further filtered out.

\begin{table}[t]
\begin{center}
\footnotesize
\renewcommand{\arraystretch}{1.5}
\resizebox{\linewidth}{!}{\large
\begin{tabular}{p{3.8cm} >{\centering\arraybackslash}p{0.8cm} >{\centering\arraybackslash}p{0.8cm} >{\centering\arraybackslash}p{0.8cm}
>{\centering\arraybackslash}p{0.8cm} >{\centering\arraybackslash}p{0.8cm}
>{\centering\arraybackslash}p{0.8cm}}
\toprule
 & 0506 & 0708 & 0910 & 1112 & 0102 & 0304\\ 
\Xhline{2\arrayrulewidth}
\textbf{Soccer} \\
\midrule
Documents & 2346 & 2658 & 2803 & 2560 & 3081 & 2143 \\
Mentions & 8255 & 8817 & 9746 & 9031 & 10284 & 7541 \\
Unique expressions & 2786 & 3202 & 3231 & 2960 & 3237 & 2694 \\
Emerging expressions & - & 2320 & 1947 & 1525 & 1636 & 1148 \\
QA pairs & 1015 & 945 & 888 & 603 & 616 & 444 \\
\Xhline{2\arrayrulewidth}
\textbf{Baseball} \\
\midrule
Documents & 672 & 894 & 822 & 290 & 448 & 984 \\
Mentions & 1725 & 3279 & 3254 & 903 & 1073 & 3501 \\
Unique expressions & 813 & 1409 & 1306 & 529 & 747 & 1827 \\
Emerging expressions & - & 1071 & 848 & 255 & 375 & 980 \\
QA pairs & 734 & 911 & 730 & 194 & 320 & 800 \\
\bottomrule
\end{tabular}
}
\caption{Statistics of \textsc{DynamicER}. Each column represents a two-month period within the dataset. Unique expressions denote distinct surface forms in each month, while emerging expressions denote distinct surface forms that appear for the first time in each month.
Refer to Figure~\ref{fig:mention_variation} for the number of distinct mentions for each of the Top-50 entities.}
\label{table:statistics}
\end{center}
\end{table}

\begin{figure*}[t]
    \centering
    \includegraphics[width=1.0\textwidth]{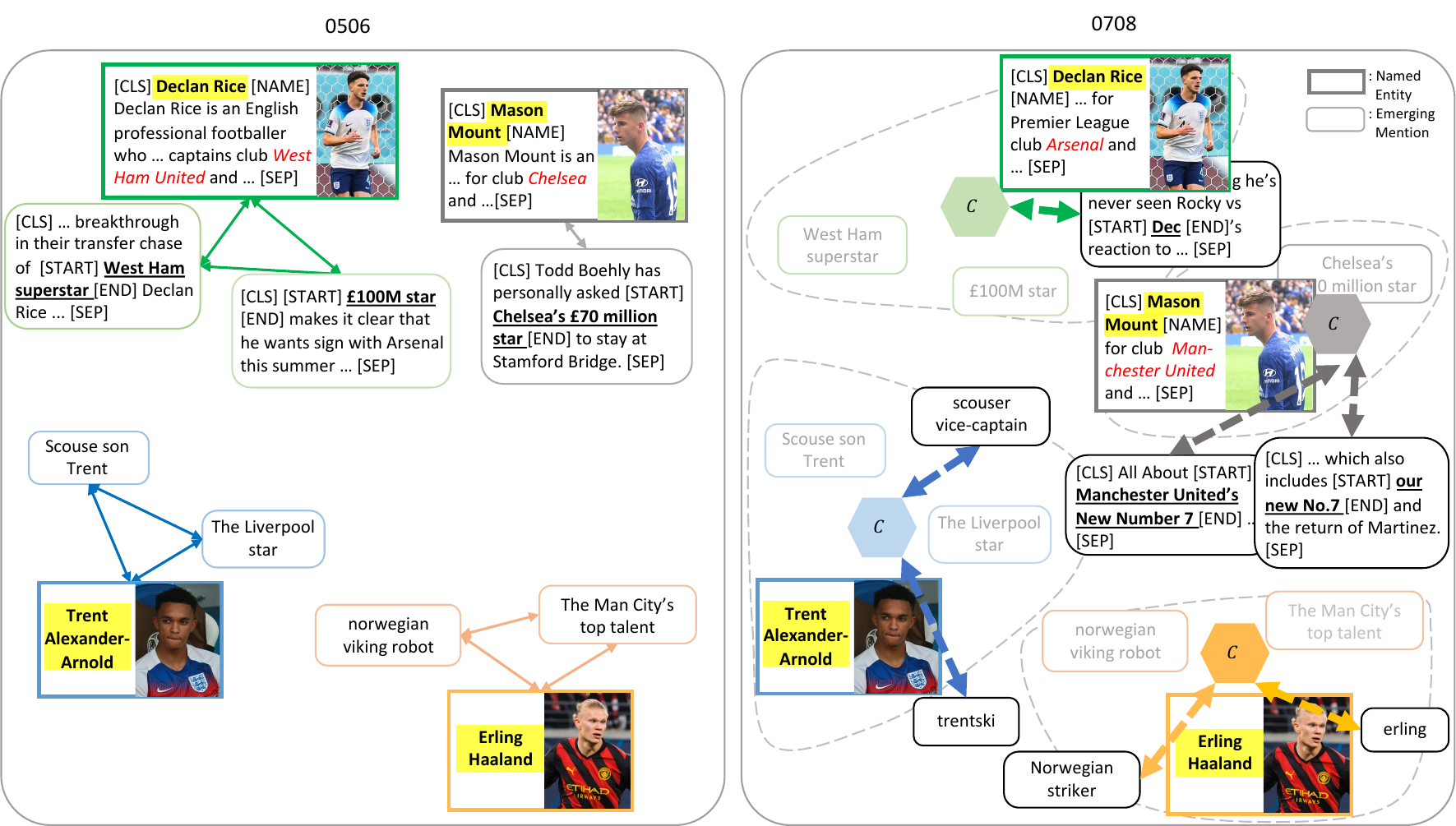}
    \caption{
    An illustrative example of TempCCA.
    $C$ denotes the representation of entity clusters formed in the previous time step. The rectangular boxes contain the entity input tokens, and the rounded boxes contain the input tokens for mention context. Entity names are highlighted, and mentions are underlined. TempCCA uses resolved mentions from the previous time step to form clusters, utilizing these cluster representations to resolve mentions in the subsequent time step. The attributes of entities that have changed are depicted in red text.    
    }
    \label{fig:figure2}
\end{figure*}

\subsection{Entity-Centric QA Pairs}
\label{subsec:entity_centric_qa}

Based on the previous annotation, we create entity-centric QA pairs, which require entity resolution to provide accurate answers.
Our basic idea is to replace each entity name in the questions with its various mentions. However, this can introduce ambiguity, as some mentions may not clearly identify the entity without additional context. For example, mentions like \textit{The Bronx Bombers} are unambiguous and can be identified as \textproc{New York Yankees} without context, whereas mentions like \textit{the winning team} are not explicit without context. Hence, before creating QA pairs, we filter out ambiguous cases that are checked by the prompt to GPT-4:
`\textit{Select the mentions from the list that
unambiguously refer to \{entity\} without
context.}'.
We further perform human verification for the remaining mentions.

Finally, we use the Wikipedia description of each entity to generate the knowledge-intensive QA pairs. To evaluate temporal challenges, we use Wikipedia articles from the revision that corresponds to the time when the mentions first appeared. With the description, we prompt GPT-4:
`\textit{Below is the description of \{entity\}.
Please generate a question-answer pair
regarding \{entity\}. The entity name itself
should be included.}'.
We instruct GPT-4 to enclose the entity within the bracket in the question text.
Once QA pairs are generated, we replace the bracketed entity name in each question with its varied mention.
The generated QA pairs are then subjected to human validation to ensure the accuracy of the question-answer pair. The details of generating QA pairs are provided in Appendix~\ref{ex:qa_generation_details}.
Finally, Table~\ref{table:statistics} shows the statistics of the labeled dataset.
\section{Approach}

Resolving new expressions based solely on the document where they first appear can be challenging due to their low lexical similarity to the entity name and the ambiguity of context. 
Thus, it can be advantageous to consider multiple documents that share similar contexts and expressions to resolve these mentions jointly. 
Previous works have studied this joint clustering approach \citep{ganea-hofmann-2017-deep, le-titov-2018-improving, angell-etal-2021-clustering, agarwal-etal-2022-entity}, but they assume a static scenario and resolve the mentions without considering the time dimension.
In our problem, on the other hand, entities evolve over time, with changes in definitions or attributes such as status, role, affiliation, or characteristics.
As exemplified in Figure~\ref{fig:figure1}, it is hard to find the coreference between ``\textit{The Angels' superstar}'' and ``\textit{The Dodgers' number 17}'' for Shohei Ohtani.
Instead, clustering mentions that appear at similar time steps could be feasible since they share events or contexts at similar time steps. Therefore, we propose a temporal segmented clustering approach with continuous adaptation (TempCCA), as shown in Figure~\ref{fig:figure2}. Our approach follows the joint clustering methods as prior works but continuously clusters the emerging mentions at each time step and further utilizes the cluster representation to resolve mentions in the next time step.

\subsection{Dual Encoder Clustering}

We follow the dual encoder clustering approach from \citet{agarwal-etal-2022-entity}.
We construct a weighted graph $G$ where the nodes represent the combined set of entities $\mathcal{E}$ and mentions $\mathcal{M}$. We then cluster these nodes based on the affinity between each pair of nodes. 
The weight of each edge is defined by affinity functions, $\phi$ and $\psi$; the former measures affinity between an entity and a mention and the latter is between mentions. For $e\in\mathcal{E}$ and $m_i, m_j\in\mathcal{M}$, we define the weight $w_{e, m_i} = -\phi(e, m_i)$ and $w_{m_i, m_j} = -\psi(m_i, m_j)$.
Each affinity function is formulated by the inner product of corresponding node embeddings:
\begin{align}
\phi(e, m_i) &= \mathbf{u}_{\mathrm{C}}(e)^\top \mathbf{u}_{\mathrm{M}}(m_i), \\
\psi(m_i, m_j) &= \mathbf{u}_{\mathrm{M}}(m_i)^\top \mathbf{u}_{\mathrm{M}}(m_j)
\end{align}
where $\mathbf{u}_{\mathrm{C}}(e)$ denotes the embedding of entity cluster formulated in the last previous time step, and $\mathbf{u}_{\mathrm{M}}(m_i)$ denotes the mention representation. The difference between \citet{agarwal-etal-2022-entity} and our work is that we formulate an entity cluster embedding, rather than using the pure output from the entity encoder.

For the entity encoder, the input tokens are structured as follows:
$\texttt{[CLS]} \, e_n \, \texttt{[NAME]} \, e_d \, \texttt{[SEP]}$,
where $e_n$ is the name of entity, and $e_d$ is the description of entity. We adopt a special token $\texttt{[NAME]}$ to separate the name and description of the entity.
For the mention encoder, the input tokens are structured as follows: 
$\texttt{[CLS]} \, c_{l} \, \texttt{[START]} \, m_i \, \texttt{[END]} \, c_{r} \, \texttt{[SEP]}$, where $c_{l}$, $c_{r}$ refer to the context to the left and right of the mention $m_i$ within the document.
We adopt special tokens $\texttt{[START]}$, and $\texttt{[END]}$ to indicate the mention span. The mention representation is defined by the output of the mention encoder, regardless of the time step.

\subsection{Continuous Training}

At the initial time step, each entity forms a single cluster.
The representation of a single cluster is simply defined by the output of the entity encoder. 
Using the obtained representations at the initial time step, we train the affinity function and their affiliated encoders.
We adopt an arborescence-based clustering approach \citep{agarwal-etal-2022-entity}. Training objectives, including positive and negative sampling, are shown in Appendix~\ref{ex:training}.

At each subsequent time step, we utilize the most recent previously resolved mentions to form an entity cluster representation:

\begin{align}
\mathbf{u}_{\mathrm{C}}(e) &= \alpha \ \mathbf{Enc}_{\mathrm{E}}(e) \nonumber \\
& + (1-\alpha) \ \frac{1}{|\mathcal{C}(e)|} \sum_{m_{i} \in \mathcal{C}(e)}\mathbf{Enc}_{\mathrm{M}}(m_{i}), \nonumber 
\end{align}
where $\mathbf{Enc}_{\mathrm{E}}(e)$ ($\mathbf{Enc}_{\mathrm{M}}(m_i)$) denotes the output of the entity (mention) encoder for the input token from entity $e$ (mention $m_i$).
$\mathcal{C}(e)$ represents mentions linked to entity $e$ in the previous time step. If the previous time step is the training phase, we use gold linking. If it is the test phase, we formulate $\mathcal{C}(e)$ with predicted linking.
The hyperparameter $\alpha$ is set to optimize the affinity models.

\section{Experiments}

\begin{table*}[ht]
\footnotesize
\begin{center}
\renewcommand{\arraystretch}{1.2}
\resizebox{\linewidth}{!}{
\begin{tabular}{p{2.4cm} >{\centering\arraybackslash}p{0.6cm} >{\centering\arraybackslash}p{0.6cm} >{\centering\arraybackslash}p{0.6cm} | >{\centering\arraybackslash}p{0.6cm} >{\centering\arraybackslash}p{0.6cm} >{\centering\arraybackslash}p{0.6cm} | >{\centering\arraybackslash}p{0.6cm} >{\centering\arraybackslash}p{0.6cm} >{\centering\arraybackslash}p{0.6cm} | >{\centering\arraybackslash}p{0.6cm} >{\centering\arraybackslash}p{0.6cm} >{\centering\arraybackslash}p{0.6cm} | >{\centering\arraybackslash}p{0.6cm} >{\centering\arraybackslash}p{0.6cm} >{\centering\arraybackslash}p{0.6cm} | >{\centering\arraybackslash}p{0.6cm} >{\centering\arraybackslash}p{0.6cm} >{\centering\arraybackslash}p{0.6cm}}
\toprule
\multirow{2}{*}{\textbf{Method}} & \multicolumn{3}{c}{Set 1 (0.0 - 0.2)} & \multicolumn{3}{c}{Set 2 (0.2 - 0.4)} & \multicolumn{3}{c}{Set 3 (0.4 - 0.6)} & \multicolumn{3}{c}{Set 4 (0.6 - 0.8)} & \multicolumn{3}{c}{Set 5 (0.8 - 1.0)} & \multicolumn{3}{c}{Total}\\
\cmidrule{2-19}
& 1112 & 0102 & 0304 & 1112 & 0102 & 0304 & 1112 & 0102 & 0304 & 1112 & 0102 & 0304 & 1112 & 0102 & 0304 & 1112 & 0102 & 0304\\
\midrule
SpEL & 56.72 & 32.69 & 32.65 & 60.34 & 55.67 & 58.82 & 63.57 & 62.68 & 64.58 & 78.66 & 81.12 & 81.02 & 78.80 & 77.57 & 76.28 & 73.28 & 72.10 & 70.73 \\
c-SpEL & 59.70 & 31.73 & 33.67 & 58.65 & 56.20 & 59.05 & 63.57 & 60.35 & 65.05 & 78.17 & 80.56 & 80.08 & 80.41 & 75.69 & 75.03 & 73.60 & 70.97 & 70.24 \\
ArboEL & 60.33 & 50.89 & 54.95 & 78.18 & 73.16 & 72.55 & 84.78 & 80.21 & 82.59 & \textbf{91.36} & \textbf{90.11} & 90.52 & 92.89 & 91.66 & 95.02 & 87.67 & 84.67 & 86.03\\
TempCCA (Ours) & \textbf{68.60} & \textbf{52.31} & \textbf{58.42} & \textbf{80.00} & \textbf{75.03} & \textbf{75.34} & \textbf{85.73} & \textbf{83.30} & \textbf{83.61} & 90.96 & 90.00 & \textbf{90.62} & \textbf{94.73} & \textbf{93.19} & \textbf{95.51} & \textbf{88.96} & \textbf{86.19} & \textbf{87.00}\\
\bottomrule
\end{tabular}
}
\caption{Results of the entity linking task by lexical similarity and time segment.}
\label{tab:exp_entitylinking_soccer}
\end{center}
\end{table*}

\begin{table}[ht]
\begin{center}
\footnotesize
\renewcommand{\arraystretch}{1.5}
\resizebox{\linewidth}{!}{\large
\begin{tabular}{p{1.7cm} >{\centering\arraybackslash}p{0.9cm} >{\centering\arraybackslash}p{0.9cm} >{\centering\arraybackslash}p{0.9cm}
>{\centering\arraybackslash}p{0.9cm} >{\centering\arraybackslash}p{0.9cm}
>{\centering\arraybackslash}p{0.9cm}}
\toprule
 & Set 1 & Set 2 & Set 3 & Set 4 & Set 5 & Total \\ 
\midrule
1112 & 242 & 1430 & 1829 & 2191 & 3339 & 9031 \\
0102 & 281 & 1658 & 2246 & 2559 & 3540 & 10284 \\
0304 & 202 & 1265 & 1769 & 2058 & 2247 & 7541 \\
\bottomrule
\end{tabular}
}
\caption{The number of mentions for each bin of lexical similarity}
\label{table:lexical_statistics}
\end{center}
\end{table}

We investigate the following research questions.
\begin{enumerate}
    \item How well does our method resolve emerging mentions compared to existing entity linking and coreference methods?
    \item Can resolving new mentions assist the RAG model in a knowledge-intensive task? 
    \item In which cases does this resolution contribute to its generative capabilities?
\end{enumerate}

\subsection{Experimental Setup}

\subsubsection{Entity Linking Task}
\textbf{Baselines.} 
We use the following models as baselines: (i) \textbf{SpEL} \citep{shavarani-sarkar-2023-spel}: the structured prediction entity linking approach, achieving the state-of-the-art performance on the AIDA-CoNLL Dataset \citep{hoffart2011robust}, (ii) \textbf{c-SpEL}: continuously trained SpEL over each time segment, (iii) \textbf{ArboEL} \citep{agarwal-etal-2022-entity}: the state-of-the-art model on MedMentions \citep{mohan2018medmentions}, and (iv) \textbf{TempCCA}: our temporal segment clustering approach with continuous adaptation. We use the dual-encoder setting from ArboEL for fair evaluation.

\textbf{Training and Inference.}
We divide our entity linking dataset into disjoint training and test time steps in timeline; specifically, documents dated from 2023-05-01 to 2023-10-31 constitute the training time steps, while documents from 2023-11-01 to 2024-04-30 form the test time steps.
For each training time step, we further split the data into training and validation sets.
For soccer, across the time steps 0506, 0708, and 0910, we use 6681, 7042, and 7783 mentions for training, and 1574, 1775, and 1963 mentions for validation, respectively.
For baseball, we use 1417, 2648, and 2592 mentions for training, and 308, 631, 662 mentions for validation, respectively.

TempCCA and c-SpEL undergo continuous training and inference. 
Specifically, we divide the documents into two-month intervals, resulting in three cycles (0506, 0708, 0910) of continuous training and three cycles (1112, 0102, 0304) of continuous inference.

\subsubsection{Entity-Centric QA}
\textbf{Baselines}.
We use four types of baselines for the entity-centric QA: (i) \textbf{LLM} (e.g.,  Llama3-8B-Instruct), (ii) \textbf{LLM-ER}: LLM with the top-1 entity linking prediction, (iii) \textbf{RaLM}~\citep{ram2023context}: LLM with concatenated top-\textit{k} retrievals, (iv) \textbf{RaLM-CoT}: RaLM with a prompt similar to zero-shot Chain-of-Thought \citep{kojima2022large}, where we first ask the LLM to resolve the mention in the question to an entity and then answer the question, and (v) \textbf{RaLM-ER}: RaLM with the top-1 entity linking prediction. We use the E5 \citep{wang2022text} as the retriever and Llama3-8B-Instruct as the generator (see Llama3 Documentation), which are state-of-the-art models. 

For LLM-ER and RaLM-ER, we utilize TempCCA to perform entity linking to resolve target mentions in question and then provide the top-1 entity prediction in the LLM's prompt. To provide TempCCA's top-1 entity linking prediction in the prompt for LLM-ER and RaLM-ER, we insert a sentence `\textit{The \{mention\} may also be referred to as \{top-1 entity prediction\}.}' right before the question. Additionally, we provide the LLM with top-3 retrievals for all RAG baselines using the format `\textit{Context: \{concatenated retrievals\}}' in the beginning. The exact format for each baseline can be found in the Appendix~\ref{ex:exDetails}.

\textbf{RAG}. Embedding all Wikipedia documents in the database requires significant computation, so we randomly select 100K articles, including the articles used for dataset collection. We create separate databases for each genre.
For soccer, we select articles linked to \textit{Category:Association football}, while for baseball, we select from \textit{Baseball}, \textit{Basketball}, and \textit{American football} to ensure enough articles.
We parse each article using the LangChain document loader (see LangChain Documentation), and index the documents using FAISS \citep{johnson2019billion} following \citet{shi2023replug}. We chunk the documents with a maximum of 1500 characters, ensuring a 10-character overlap between chunks.

\textbf{Metrics}. To evaluate generated answers, we use the F1 score following \citet{petroni2021kilt}. Since most answers in our QA dataset are either a noun phrase or a short sentence, we only consider the first sentence of each answer. We parse this first sentence using the nltk sentence tokenizer.\footnote{https://www.nltk.org/api/nltk.tokenize.sent\_tokenize.html}

\subsection{Experimental Results}

We present the performance of our entity linking and entity-centric QA tasks in the soccer genre. Additionally, the performance in the baseball genre is reported in Appendix~\ref{ex:additionalExperiments}.

\subsubsection{Results of Entity Linking}

Table~\ref{tab:exp_entitylinking_soccer} presents the accuracy of our entity linking task in the soccer genre. 
To rigorously evaluate the performance of each method in resolving mentions across different levels of lexical similarity, we present the results for each bin of lexical similarity separately. For each mention, we calculate the Jaccard similarity \citep{zhang2021semantic} with the entity name using a character-level token as a straightforward measure of lexical overlap. 
The number of mentions for each bin is presented in Table~\ref{table:lexical_statistics}.

The results reveal a clear trend: as Jaccard similarity increases, the accuracy of all baselines improves. 
This underscores the importance of lexical similarity in entity linking tasks, where emerging mentions with lower lexical similarity typically lead to less accurate linking of mentions.

c-SpEL surpasses SpEL in 1112, but falls below in 0102 and 0304 in total. This implies that the performance of c-SpEL deteriorates as it moves further from the last time point of learning.
Besides, TempCCA consistently outperforms other baselines in most cases, except for Set 4 (0.6 - 0.8), when ArboEL exceeds TempCCA from 0.11 to 0.4. Nevertheless, TempCCA shows a robust performance across all months and sets.
Notably, TempCCA surpasses the baselines the most in Set 1, from 1.42 in 0102 to 8.27 in 1112.
This implies that utilizing the recently predicted mentions can help jointly resolve mentions with low lexical overlap, as there tend to be similar mentions within a similar time step.

\subsubsection{Results of Entity-Centric QA}

Table~\ref{tab:qa_result} presents the performance of our entity-centric QA task in the soccer genre.
The accuracy of TempCCA's top-1 entity linking prediction on QA questions attains 66.62, 67.62, and 65.86 for 1112, 0102, and 0304, respectively.
Compared to the base LLM, LLM-ER improves performance by an average of 3, indicating that resolving new mentions helps the LLM generate more accurate responses. Still, both LLM and LLM-ER underperform the RAG baselines. Specifically, RaLM improves the LLM's performance by an average of 15 using the retrieved documents.
Interestingly, RaLM-CoT performs significantly worse than RaLM by an average of 6.8, suggesting that entity prediction by the LLM itself does not effectively contribute to QA accuracy. On the other hand, RaLM-ER improves RaLM by an average of 1.4 and outperforms all other baselines, indicating that resolving mentions can be beneficial to knowledge-intensive QA tasks.

\begin{table}[t]
\begin{center}
\renewcommand{\arraystretch}{1.05}
\resizebox{\linewidth}{!}{
{\footnotesize
\begin{tabular}{p{2.7cm} >{\centering\arraybackslash}p{1.2cm} >{\centering\arraybackslash}p{1.2cm} >{\centering\arraybackslash}p{1.2cm}}
\toprule
 & 1112 & 0102 & 0304 \\  
\midrule
Average \\
\midrule
    LLM & 28.82 & 27.84 & 29.75 \\
    LLM-ER (Ours) & 31.47 & 32.02 & 32.15 \\
    RaLM & 44.48 & 42.75 & 46.55 \\
    RaLM-CoT & 38.09 & 36.43 & 38.56 \\
    RaLM-ER (Ours) & \textbf{45.67} & \textbf{44.60} & \textbf{47.93} \\
\midrule
\multicolumn{2}{l}{Retrieval Hit} \\
\midrule
    LLM & 28.48 & 28.00 & 31.33 \\
    LLM-ER (Ours) & 32.91 & 33.61 & 34.74 \\
    RaLM & 59.07 & 56.37 & 59.31 \\
    RaLM-CoT & 50.55 & 48.01 & 48.03 \\
    RaLM-ER (Ours) & \textbf{59.42} & \textbf{56.71} & \textbf{60.24} \\
\midrule
\multicolumn{2}{l}{Retrieval Miss} \\
\midrule
    LLM & 29.23 & 27.64 & 27.42 \\
    LLM-ER (Ours) & \textbf{29.69} & \textbf{30.09} & 28.33 \\
    RaLM & 26.60 & 26.30 & 27.74 \\
    RaLM-CoT & 22.82 & 22.45 & 24.59 \\
    RaLM-ER (Ours) & 28.84 & 29.98 & \textbf{29.76} \\
\bottomrule
\end{tabular}
}
}
\caption{Results of entity-centric QA for each time segment in F1 scores. Our approach is applied to both LLM and RAG framework.}
\label{tab:qa_result}
\end{center}
\end{table}

\begin{table}[t]
\begin{center}
\renewcommand{\arraystretch}{1.2}
\resizebox{\linewidth}{!}{
{\footnotesize
\begin{tabular}{p{2.4cm} >{\centering\arraybackslash}p{0.8cm} >{\centering\arraybackslash}p{0.8cm} >{\centering\arraybackslash}p{0.8cm} >{\centering\arraybackslash}p{0.8cm}}
\toprule
Retrieval & \multicolumn{2}{c}{Hit} & \multicolumn{2}{c}{Miss} \\  
Entity Linking & Success & Failure & Success & Failure\\ 
\midrule
    RaLM & 58.78 & \textbf{60.43} & 29.86 & 23.42  \\
    RaLM-ER (Ours) & \textbf{60.01} & 59.48 & \textbf{32.12} & \textbf{25.29}\\
\bottomrule
\end{tabular}
}
}
\caption{Comparison of RaLM and RaLM-ER performance in retrieval hits and misses, and entity linking successes and failures.}
\label{tab:qa_result_r_el}

\end{center}
\end{table}

To analyze the results comprehensively, we report the results separately for cases of retrieval success (retrieval hit) and failure (retrieval miss). As shown in the middle section of Table~\ref{tab:qa_result}, RAG baselines significantly enhance LLM performance; RaLM improves the base LLM by 30.73, 28.37, and 27.98 in 1112, 0102, and 0304, respectively. Additionally, RaLM-ER performs on par with or slightly better than RaLM when the retrieval succeeds, despite potential errors in entity resolution. Conversely, when retrieval fails, RaLM performs worse than the base LLM, with a decrease of 3.16 in 1112. This highlights the critical impact of retrieval failures. On the other hand, LLM-ER enhances the base LLM in all cases. 
Furthermore, RaLM-ER mitigates hallucinations, improving RaLM by approximately 2.06, 3.58, and 2.02 in 1112, 0102, and 0304, respectively. Remarkably, RaLM-ER even outperforms all baselines despite incorrect retrievals in 0304.

To further pinpoint the improvement, we analyze the results of RaLM and RaLM-ER in four scenarios: when entity resolution is correct or incorrect, within the context of both retrieval hits and misses. Table~\ref{tab:qa_result_r_el} shows the averaged results across time segments. When retrieval is successful, the performance of RaLM-ER improves by 1.2 when the entity resolution is correct; however, it drops approximately 1.0 when the entity resolution is wrong. Conversely, when retrieval fails, performance is enhanced in both correct and incorrect resolution cases. RaLM-ER improves upon RaLM by 2.3 when the resolution is correct and by 1.8 when the resolution is incorrect. Although it seems crucial to avoid introducing incorrect resolutions, providing entity resolution results to the LLM generally improves end-to-end performance. 

\section{Conclusion}

In this work, we addressed the challenge of resolving new linguistic expressions in the dynamic and ever-evolving landscape of human language. We introduced \textproc{DynamicER} to evaluate the ability of models to resolve emerging mentions. Our benchmark proposes entity linking tasks for resolving emerging mentions and entity-centric QA tasks for RAG evaluation. 
To address the temporal dynamics of emerging mentions, we proposed a temporal segmented clustering method with continual adaptation.
Our exhaustive experiments demonstrated that our method surpassed existing baselines in resolving new expressions, particularly when there is less lexical overlap. 

Future work may extend to updating KB using entity resolution, which can directly handle the retrieval failures caused by mention dynamics. 
Through \textproc{DynamicER}, we provide a resource for the research community to further explore and improve dynamic entity resolution. We hope this fosters further research towards developing more robust models capable of handling the continuous emergence of new expressions.
\section*{Limitations}

We acknowledge several limitations in our work. Firstly, our dataset and method are primarily designed to handle variations in single entity mentions and may not effectively address cases where multiple entities are combined into a single mention, such as "Kimye" referring to Kanye West and Kim Kardashian. Future research could explore developing benchmarks and models capable of resolving such combined mentions into multiple entities.
Additionally, our dataset may reflect biases introduced by the GPT-4 model and the specific prompts used during its creation. Although we perform thorough human validation and revisions, future studies could benefit from employing a diverse set of language models to mitigate these potential biases.

\section*{Ethics Statement}

\textbf{Safety}.
All data were sourced from Tumblr, which is publicly available.
To ensure the safety of our dataset, we conduct a two-stage filtering process. Initially, annotators were instructed to report any potentially harmful or privacy-invading content. Following this, the authors reviewed the remaining content to further filter out inappropriate materials. Despite these efforts, biases such as stereotyping may still be present due to the nature of real communication on Tumblr, where a significant portion of the user base consists of teenagers and young adults.

\textbf{Intended Use}.
The \textsc{DynamicER} dataset is intended to be used for research purposes only, and the use is subject to Tumblr Terms of Service and Community Guidelines.

\textbf{Annotator Compensation}.
We hired university students as annotators. To uphold ethical standards, we compensated our annotators with a fair hourly wage of approximately USD \$15. The estimated completion time for each task was determined through multiple preliminary trials conducted by our research team. Consequently, the average expense per datapoint amounted to approximately \$0.30. Data points requiring additional time were compensated at a proportionately higher rate to ensure fairness. The full text of instructions is provided in Figure~\ref{tab:fulltext1}.

\section*{Acknowledgements}
We thank Wonkwang Lee, Chris Dongjoo Kim, Sangwoo Moon, Sehun Lee, Jongchan Noh, and the anonymous reviewers for their insightful discussions. This work was supported by 
Institute of Information \& communications Technology Planning \& Evaluation (IITP) grant funded by the Korea government (MSIT) (No.~RS-2022-II220156, Fundamental research on continual meta-learning for quality enhancement of casual videos and their 3D metaverse transformation), 
the SNU-Global Excellence Research Center establishment project, 
the National Research Foundation of Korea (NRF) grant funded by the Korea government (MSIT) (No.~2023R1A2C2005573),
Basic Science Research Program through the National Research Foundation of Korea(NRF) funded by the Ministry of Education(RS-2023-00274280), 
and Institute of Information \& communications Technology Planning \& Evaluation (IITP) grant funded by the Korea government (MSIT) (No.~RS-2021-II211343, Artificial Intelligence Graduate School Program (Seoul National University)). Gunhee Kim is the corresponding author.

\bibliography{custom}

\appendix
\newpage
\section{Experimental Details}\label{ex:exDetails}

\textbf{Dataset generation}. When we use GPT to generate dataset, we use gpt-4-turbo and set the max\_token as 1024 and the temperature 0. 

\textbf{EL Baseline Training}. 
For SpEL, we use \texttt{roberta-base} for the encoder and conduct only the final fine-tuning step based on the second-step pretrained model. We train the model for 10 epochs while setting the batch\_size to 16, bert\_dropout to 0.2, and label\_size to 10240. We report the results of the micro entity linking metrics.
For ArboEL, we set the batch\_size to 32 and use the other hyperparameters the same as in the ArboEL setting without any modifications.

\textbf{TempCCA}. For each training time step, we train the bi-encoder (\texttt{bert-base-cased}) \citep{DBLP:journals/corr/abs-1810-04805} model for 5 epochs with a batch size of 32 and a learning rate of 3e-05 using the Adam optimizer. In each training iteration, we randomly select 30 mentions to form an entity cluster if the number of previously resolved mentions for that entity exceeds 30. We set the hyperparameter $\alpha$ as 0.8.
It requires A6000 X 4 GPUs for training and takes 6 hours for continual training.

\textbf{RAG}. We use e5-base for the retriever from hugging face \texttt{intfloat/e5-base}. We set \textit{temperature} and \textit{max\_new\_tokens} to 0.3 and 30, respectively, for all baselines except the first text generation in RaLM-CoT, for which we set \textit{temperature} to 0.1 and \textit{new\_tokens} to 10 to ensure accurate entity prediction. 
\newpage
\textbf{QA}. The prompt formats for five different QA baselines are presented below. 
\begin{table}[H]
\renewcommand{\arraystretch}{1.1}
\resizebox{\linewidth}{!}{\footnotesize
\begin{tabular}{!{\vrule width 0.5pt} p{\linewidth} !{\vrule width 0.5pt}}
    \Xhline{0.5pt}
    \# LLM\\
    Given a question, please provide a short answer.  \\
    Question: \{question\}\\
    Answer:\\
    \\
    \# LLM-ER\\
    The mention \{mention\} may also be referred to as \{entity\}. Given a question, please provide a short answer.\\
    Question: \{question\}\\
    Answer:\\
    \\
    \# RaLM\\
    Context: \{context\}\\
    Given a question, please provide a short answer.  \\
    Question: \{question\}\\
    Answer:\\
    \\
    \# RaLM-CoT\\
    \#\# first\\
    Context: \{context\}\\
    Question: \{question\} \{mention\} is\\
    \#\# second\\
    Context: \{context\}\\
    Question: \{quesition\} \{mention\} is \{first answer\}.\\
    Answer:\\
    \\
    \# RaLM-ER\\
    Context: \{context\}\\
    The mention \{mention\} may also be referred to as \{entity\}. Given a question, please provide a short answer.  \\
    Question: \{question\}\\
    Answer:\\
    \Xhline{0.5pt}
\end{tabular}
}
\caption{Prompt template for QA task}
\label{tab:qa_prompt}
\end{table}

\newpage
\section{Additional Experimental Results}\label{ex:additionalExperiments}

\textbf{Results for the Baseball Genre}. We report the performance of our entity linking and entity-centric QA task in the baseball genre. As shown in Table~\ref{tab:exp_entitylinking_baseball}, TempCCA demonstrates robust performance across all months. TempCCA exceeds ArboEL from 2.14 in 0102 to 3.54 in 0304.
In Table~\ref{tab:qa_result_baseball}, LLM-ER exceeds the base LLM by an average of 1.6. RaLM consistently enhances the LLM's performance by an average of 16 using the retrieved documents. RaLM-ER improves RaLM by an average of 0.8 and outperforms all other baselines.
In cases of retrieval success (retrieval hit), the RAG baselines significantly boost LLM performance. In this case, RaLM-ER demonstrates the best performance. On the contrary, when retrieval fails, RaLM performs worse than the base LLM. However, LLM-ER still improves upon the base LLM in these cases. RaLM-ER effectively mitigates hallucinations, improving RaLM by 1.14, 0.85, and 1.12 in 1112, 0102, and 0304, respectively.

\begin{table}[H]
\begin{center}
\renewcommand{\arraystretch}{1.05}
\resizebox{\linewidth}{!}{
{\footnotesize
\begin{tabular}{p{2.7cm} >{\centering\arraybackslash}p{1.2cm} >{\centering\arraybackslash}p{1.2cm} >{\centering\arraybackslash}p{1.2cm}}
\toprule
\textbf{Method} & 1112 & 0102 & 0304 \\  
\midrule
    SpEL & 59.93 & 53.99 & 67.80 \\
    c-SpEL & 57.45 & 53.07 & 66.59 \\
    ArboEL & 85.38 & 85.93 & 85.66 \\
    TempCCA & \textbf{88.26} & \textbf{88.07} & \textbf{89.20} \\
\bottomrule
\end{tabular}
}
}
\caption{Results of the entity linking task by lexical similarity and time segment.}
\label{tab:exp_entitylinking_baseball}
\end{center}
\end{table}

\begin{table}[H]
\begin{center}
\renewcommand{\arraystretch}{1.05}
\resizebox{\linewidth}{!}{
{\footnotesize
\begin{tabular}{p{2.7cm} >{\centering\arraybackslash}p{1.2cm} >{\centering\arraybackslash}p{1.2cm} >{\centering\arraybackslash}p{1.2cm}}
\toprule
 & 1112 & 0102 & 0304 \\  
\midrule
Average \\
\midrule
    LLM & 30.25 & 31.92 & 28.38 \\
    LLM-ER (Ours) & 31.59 & 34.54 & 29.16 \\
    RaLM & 44.54 & 50.45 & 45.06 \\
    RaLM-CoT & 36.92 & 42.76 & 38.77 \\
    RaLM-ER (Ours) & \textbf{45.56} & \textbf{51.13} & \textbf{45.70} \\
\midrule
\multicolumn{2}{l}{Retrieval Hit} \\
\midrule
    LLM & 30.05 & 30.65 & 30.38 \\
    LLM-ER (Ours) & 32.64 & 33.84 & 31.37 \\
    RaLM & 63.27 & 63.18 & 62.34 \\
    RaLM-CoT & 50.94 & 53.14 & 51.89 \\
    RaLM-ER (Ours) & \textbf{64.15} & \textbf{63.73} & \textbf{62.58} \\
\midrule
\multicolumn{2}{l}{Retrieval Miss} \\
\midrule
    LLM & 30.44 & 33.62 & 25.94 \\
    LLM-ER (Ours) & \textbf{30.63} & \textbf{35.48} & \textbf{26.46} \\
    RaLM & 27.30 & 33.44 & 23.95 \\
    RaLM-CoT & 24.01 & 28.90 & 22.74 \\
    RaLM-ER (Ours) & 28.44 & 34.29 & 25.07 \\
\bottomrule
\end{tabular}
}
}
\caption{Results of entity-centric QA for each time segment in F1 scores. Our approach is applied to both LLM and RAG framework.}
\label{tab:qa_result_baseball}
\end{center}
\end{table}

\textbf{Additional Factual Accuracy Evaluation}.
To ensure robustness on evaluating entity-centric QA tasks, we conduct an additional evaluation using GPT-4 to assess factual accuracy. Following the methodology of \citet{kamalloo-etal-2023-evaluating}, we prompt GPT-4 to assess factual accuracy and detect potential hallucinations in candidate answers using the following zero-shot prompt: `\textit{Question: \{question\}, Answer: \{gold answer\}, Candidate: \{prediction\}: Is candidate correct?}'. Table~\ref{tab:gpt4_eval} presents the results in the soccer genre. Consistent with the results of F1 scores, our LLM-ER method improves upon the standard LLM, and our RaLM-ER significantly enhances RaLM performance.

\begin{table}[H]
\begin{center}
\renewcommand{\arraystretch}{1.05}
\resizebox{\linewidth}{!}{
{\footnotesize
\begin{tabular}{p{2.7cm} >{\centering\arraybackslash}p{1.2cm} >{\centering\arraybackslash}p{1.2cm} >{\centering\arraybackslash}p{1.2cm}}
\toprule
GPT-4 score & 1112 & 0102 & 0304 \\  
\midrule
    LLM & 26.37 & 23.86 & 30.47 \\
    LLM-ER (Ours) & 26.53 & 23.70 & 30.93 \\
    RaLM & 59.70 & 62.50 & 63.43 \\
    RaLM-CoT & 57.14 & 60.95 & 57.05 \\
    RaLM-ER (Ours) & \textbf{62.85} & \textbf{63.80} & \textbf{66.59} \\
\bottomrule
\end{tabular}
}
}
\caption{Factual accuracy results for entity-centric QA tasks assessed by GPT-4.}
\label{tab:gpt4_eval}
\end{center}
\end{table}

\textbf{Analysis of Entity-Centric QA task using Jaccard similarity}.
We report that lower Jaccard similarity between mentions and entity names degrades entity linking performance. To further investigate this, we analyze the entity-centric QA task using Jaccard similarity to observe how the lexical overlap of mentions in questions affects the answer generation performance.
Table~\ref{table:jaccard_qa} shows the results in the soccer genre. Our finding indicates that as Jaccard similarity increases, the end-to-end generation accuracy also improves in the QA task. This shows that the RAG performance may degrade with variational mention surface forms. Notably, the score gap between RaLM and RaLM-ER increases as the Jaccard similarity decreases. It implies that effectively resolving emerging mentions is critical to enhance the end-to-end performance.

\begin{table}[ht]
\begin{center}
\footnotesize
\renewcommand{\arraystretch}{1.5}
\resizebox{\linewidth}{!}{\large
\begin{tabular}{p{2.0cm} >{\centering\arraybackslash}p{1.5cm} >{\centering\arraybackslash}p{1.5cm} >{\centering\arraybackslash}p{1.5cm}
>{\centering\arraybackslash}p{1.5cm} >{\centering\arraybackslash}p{1.5cm}}
\toprule
 & Set 1 \newline (0.0-0.2) & Set 2 \newline (0.2-0.4) & Set 3 \newline (0.4-0.6) & Set 4 \newline (0.6-0.8) & Set 5 \newline (0.8-1.0)\\ 
\midrule
RaLM & 29.90 & 32.91 & 43.32 & 46.04 & 47.75 \\
RaLM-ER & 34.98 & 35.46 & 44.80 & 48.20 & 48.36 \\
\bottomrule
\end{tabular}
}
\caption{Analysis of entity-centric QA task using Jaccard similarity}
\label{table:jaccard_qa}
\end{center}
\end{table}

\textbf{Candidate Generation Results}. 
We also report the Recall@\textit{n} score in the soccer genre for ArboEL and TempCCA in Table~\ref{tab:exp_candidate64}. Recall@\textit{n} score considers a prediction successful if the correct entity appears within the top-\textit{n} predicted entities. 
TempCCA consistently outperforms ArboEL regardless of top-\textit{n}, which indicates that our setting of temporal segmented clustering surpasses the baseline performance in entity candidate generation as well.

\begin{table}[H]
\small
\begin{center}
\renewcommand{\arraystretch}{1.2}
\resizebox{\linewidth}{!}{
\begin{tabular}{p{2.4cm} >{\centering\arraybackslash}p{0.5cm} >{\centering\arraybackslash}p{0.5cm} >{\centering\arraybackslash}p{0.5cm} >{\centering\arraybackslash}p{0.5cm} >{\centering\arraybackslash}p{0.5cm} >{\centering\arraybackslash}p{0.5cm} >{\centering\arraybackslash}p{0.5cm}}
\toprule
\textbf{Recall@} & 1 & 2 & 4 & 8 & 16 & 32 & 64\\
\midrule
\textbf{1112}\\
\midrule
ArboEL & 87.67 & 90.01 & 91.74 & 92.95 & 94.26 & 95.13 & 95.90 \\
TempCCA (Ours) & \textbf{88.96} & \textbf{90.80} & \textbf{92.51} & \textbf{93.62} & \textbf{94.60} & \textbf{95.46} & \textbf{96.19} \\
\midrule
\textbf{0102}\\
\midrule
ArboEL & 84.67 & 87.69 & 89.78 & 91.44 & 92.82 & 93.98 & \textbf{95.18} \\
TempCCA (Ours) & \textbf{86.19} & \textbf{88.70} & \textbf{90.71} & \textbf{92.16} & \textbf{93.31} & \textbf{94.18} & \textbf{95.18} \\
\midrule
\textbf{0304}\\
\midrule
ArboEL & 86.03 & 88.91 & 90.87 & 92.19 & 93.38 & 94.56 & 95.58 \\
TempCCA (Ours) & \textbf{87.00} & \textbf{89.54} & \textbf{91.34} & \textbf{92.77} & \textbf{93.93} & \textbf{94.91} & \textbf{95.60} \\
\bottomrule
\end{tabular}
}
\caption{Results of candidate generation}
\label{tab:exp_candidate64}
\end{center}
\end{table}

\section{Tag List}\label{ex:tagList}
Below are the tags used to scrape the documents:

Soccer genre: \#AC Milan, \#Arsenal FC, \#Atletico Madrid, \#Borussia Dortmund, \#Bundesliga, \#Chelsea FC, \#FC Barcelona, \#FC Bayern, \#Juventus, \#La Liga, \#Ligue 1, \#Liverpool FC, \#Manchester City, \#Manchester United, \#Premier League, \#PSG, \#Real Madrid, \#Serie A, \#Tottenham Hotspur, \#UCL, \#West Ham

Baseball genre: \#Arizona Diamondbacks, \#Atlanta Braves, \#Baltimore Orioles, \#Boston Red Sox, \#Chicago Cubs, \#Chicago White Sox, \#Cincinnati Reds, \#Cleveland Guardians, \#Colorado Rockies, \#Detroit Tigers, \#Houston Astros, \#Kansas City Royals, \#Los Angeles Angels, \#Los Angeles Dodgers, \#Miami Marlins, \#Milwaukee Brewers, \#Minnesota Twins, \#New York Mets, \#New York Yankees, \#Oakland Athletics, \#Philadelphia Phillies, \#Pittsburgh Pirates, \#San Diego Padres, \#San Francisco Giants, \#Seattle Mariners, \#St. Louis Cardinals, \#Tampa Bay Rays, \#Texas Rangers, \#Toronto Blue Jays, \#Washington Nationals
\section{QA Generation Details}\label{ex:qa_generation_details}

In this section, we describe the details of QA generation. We first filter out ambiguous mentions using GPT-4 as described in Section \ref{subsec:entity_centric_qa}. For the remaining mentions, we present a few examples as shown in Table \ref{tab:filterAmbiguousPrompt} and provide each mention to three annotators. They are tasked with distinguishing between ambiguous and unambiguous mentions. Only mentions judged as unambiguous by at least two annotators are retained, and the rest are filtered out.

Following the procedure of initial generation from \citet{ko2024growover}, we then generate entity-centric QA pairs. First, we gather articles related to the entity referred to by each mention. For each article, we extract a paragraph between 300 and 2000 characters in length and perform random sampling based on the number of unambiguous mentions referring to the entity. We then provide GPT-4 with the prompt shown in Table \ref{tab:generateQAprompt} to generate the QA pairs. To simultaneously evaluate the success of retrieval model in the RAG framework, we annotate the evidence text required to answer the generated questions correctly. The evidence text is the paragraph given as input to GPT-4. By following this method, we create entity-centric QA pairs, and subsequently, we replace the entity names in the questions with the corresponding mentions. After generating QA pairs, authors validate whether the generated QA pairs are answerable given questions, filtering out the pairs if not.
\section{Training Procedure}\label{ex:training}
We adopt \citet{agarwal-etal-2022-entity}'s training procedure, which shows state-of-the-art performance on the MedMentions dataset \citep{mohan2018medmentions}. In this section, we summarize \citet{agarwal-etal-2022-entity}'s procedure, including positive and negative sampling. The goal is to optimize the affinity functions by using batch-wise gradient optimization. For each incoming batch of mentions $M_B$, we construct a graph $G_{M_B}$, where the nodes represent each mention $m_i \in M_B$, mentions coreferent with $m_i$, and the that include each mention $m_i \in M_B$, mentions coreferent with $m_i$, and the corresponding set of gold entities for each mention in the batch. 
Consequently, the full set of edges in the graph $G_{M_B}$ for batch is represented as:

\begin{equation}
\begin{aligned}
E(G_{M_B}) &= \bigcup_{m_i \in M_B} \Big( \{ (e_i^*, m_k) \mid m_k \in \mathcal{S}_{e_i^*} \} \\
           & \qquad \cup \{ (m_k, m_l) \mid m_k, m_l \in \mathcal{S}_{e_i^*} \} \Big)
\end{aligned}
\end{equation}

Here, \(\mathcal{S}_{e_i^*}\) represents the set of mentions coreferent with the gold entity \(e_i^*\).

\subsection*{Positive and Negative Sampling}

The graph \(G_{M_B}\) is partitioned into disjoint clusters. The constraints for each cluster \(C\) are:

\begin{enumerate}
    \item \(C\) includes at most one entity.
    \item For all \(u, v \in C\), if \(u\) is connected to \(v\), then \(f(u, v) \leq \lambda\),
    \item For all \(u, v \in C\), either \(u\) is connected to \(v\) or \(v\) is connected to \(u\).
\end{enumerate}

Following \citet{angell-etal-2021-clustering, agarwal-etal-2022-entity}, we sort edges by decreasing dissimilarity, and iteratively remove edges. 
First, we eliminate all edges in graph \(G\) with weights exceeding \(\lambda\). Next, we process each edge \((u, v) \in E\) in descending order of dissimilarity, checking if its presence violates any of the three defined constraints. If an edge does violate a constraint, we remove it from \(E\). If not, we examine whether the connected component of node \(u\) contains an entity, i.e., \(|C_u \cap \mathcal{E}| = 1\). If it does, we drop edge \((u, v)\) if \(v\) can still be reached by an entity node without \((u, v)\). We retain edge \((u, v)\) and continue iteration if not reachable, preserving the connectivity of the cluster. Our predicted clusters are the final connected components in graph \(G\).

For negative sampling, we identify negative edges for each mention \(m_i \in B\). The negative edges include the \(k/2\) lowest-weight incoming edges from each of \(\mathcal{E} \setminus \{e_i^*\}\) and \(\mathcal{M} \setminus \mathcal{S}_{e_i^*}\).

\subsection*{Loss Function}

Following \citet{pmlr-v97-yadav19a, agarwal-etal-2022-entity}, the loss function \(\mathcal{L}(m_i)\) is defined as follows:
\begin{equation}
\begin{split}
\mathcal{L}(m_i) =& \sum_{p \in \kappa(m_i)} ( I_{p, m_i} \log(\sigma(w_{p, m_i})) \\ & + (1 - I_{p, m_i}) \log(1 - \sigma(w_{p, m_i})) )
\end{split}
\end{equation}
where \(\kappa(m_i)\) includes all neighbors with outgoing edges to \(m_i\) in the graph, \(I_{p, m_i}\) is an indicator variable, with \(I_{p, m_i} = 1\) if \((p, m_i)\) is in a pruned set of edges; otherwise, \(I_{p, m_i} = 0\). \(\sigma(\cdot)\) denotes the softmax function. The total loss for the batch \(B\) is the average of the losses for all mentions in \(B\), optimizing for higher probabilities for positive edges and lower probabilities for negative edges.

\section{Dataset Generation Prompts}\label{ex:prompt}
Table~\ref{tab:mentionDetectionPrompt} shows the prompt used for the mention detection from the corpora. Table~\ref{tab:filterAmbiguousPrompt} shows the prompt used for filtering ambiguous mentions, which is not suitable for generating entity-centric QA. Table~\ref{tab:generateQAprompt} shows the prompt used for generating entity-centric QA.

\begin{table*}
\small
\renewcommand{\arraystretch}{1.4}
\resizebox{\textwidth}{!}{
\begin{tabular}{!{\vrule width 1.2pt} p{\linewidth} !{\vrule width 1.2pt}}
    \Xhline{1.2pt}
        Please identify all expressions in the given text that explicitly name or describe a player, coach, or team. This includes direct names, nicknames, and any role-specific references (like positions or accolades) that imply a particular individual or team.\\
        \\
        Instructions:\\
         1) Provide a list of expressions (ex. [``our striker'', ``messi'', ``agent of chaos'', ...])\\
         2) The expressions can be long, as they also include noun phrases.\\
         3) Write expressions exactly as they appear in the text. Do not modify the expressions in the text, even if they contain typos. Maintain the original capitalization and spacing, as we will use string matching.\\
         4) List expressions in the order they appear in the text. If an expression appears multiple times, list it multiple times in the order of appearance.\\
         5) If the text lacks a suitable expression, provide an empty list.\\
        \\
        Be sure to follow the following format and write your answer within the list: [``Expression 1'', ``Expression 2'', ... ]\\
        \\
        Text: \{\textit{text}\}\\
    \Xhline{1.2pt}
\end{tabular}
}
\caption{Sample prompt for mention detection}
\label{tab:mentionDetectionPrompt}
\end{table*}

\begin{table*}
\small
\renewcommand{\arraystretch}{1.4}
\resizebox{\textwidth}{!}{
\begin{tabular}{!{\vrule width 1.2pt} p{\linewidth} !{\vrule width 1.2pt}}
    \Xhline{1.2pt}
        Provided List: \{\textit{mention list}\}\\
        \\
        The provided list is a compilation of mentions identified in the textual corpora. Upon human verification, these mentions refer to \{\textit{entity}\} in the text.\\
        Select the mentions from the list that unambiguously refer to  without context.\\
        Unambiguous mentions are mentions that exclusively refer to the specified entity (\{\textit{entity}\}) without requiring additional context.\\
        \\
        Examples:\\
         Unambiguous Mentions for \textproc{Manchester\_United\_F.C.}: ``Man Utd'', ``20-time English champions'', ``the Red Devils club'', ...\\
         Ambiguous Mentions for \textproc{Manchester\_United\_F.C.}: ``United'', ``chaos club'', ``English Powerhouse'', ``the First Team'', ...\\
        \\
        Instructions:\\
         1) Write mentions exactly as they appear in the provided list. Do not modify the mentions in the list, even if they contain typos. Maintain the original capitalization and spacing.\\
         2) Be sure to follow the following format and write your answer within the list: [``Mention 1'', ``Mention 2'', ... ]\\
    \Xhline{1.2pt}
\end{tabular}
}
\caption{Sample prompt for filtering ambiguous mentions}
\label{tab:filterAmbiguousPrompt}
\end{table*}

\begin{table*}
\small
\renewcommand{\arraystretch}{1.4}
\resizebox{\textwidth}{!}{
\begin{tabular}{!{\vrule width 1.2pt} p{\linewidth} !{\vrule width 1.2pt}}
    \Xhline{1.2pt}
        Generate a Q\&A pair about [\{\textit{entity}\}] based on a given context. The context will provide factual information about [\{\textit{entity}\}].\\
        Assume the person answering the question has common sense and is aware of the details and key points in the context, but the context itself is not quoted or referenced directly.\\
        \\
        Context: \{\textit{context}\}\\
        \\
        Follow these instructions to generate a Q\&A pair:\\
         1) Provide a question and an answer.\\
         2) Bracket the corresponding Entity ([\{\textit{entity}\}]) in the Question, like the sample Q\&A pair given below.\\
         3) Do NOT use phrases such as ‘according to the context’ in your question.\\
         4) Generate a SINGLE Q\&A pair.\\
         5) Provide a SHORT ANSWER.\\
        \\
        Write your Q\&A pair within curly brackets using the following format: \{Question\}\{Answer\}\\
        \\
        Sample Q\&A pair : \{Where was [Lionel\_Messi] born?\}\{Lionel Messi was born in Rosario, Argentina.\}\\
    \Xhline{1.2pt}
\end{tabular}
}
\caption{Sample prompt for generating entity-centric QA pairs}
\label{tab:generateQAprompt}
\end{table*}

\begin{figure*}[t]
    \centering
    \includegraphics[width=1.0\textwidth]{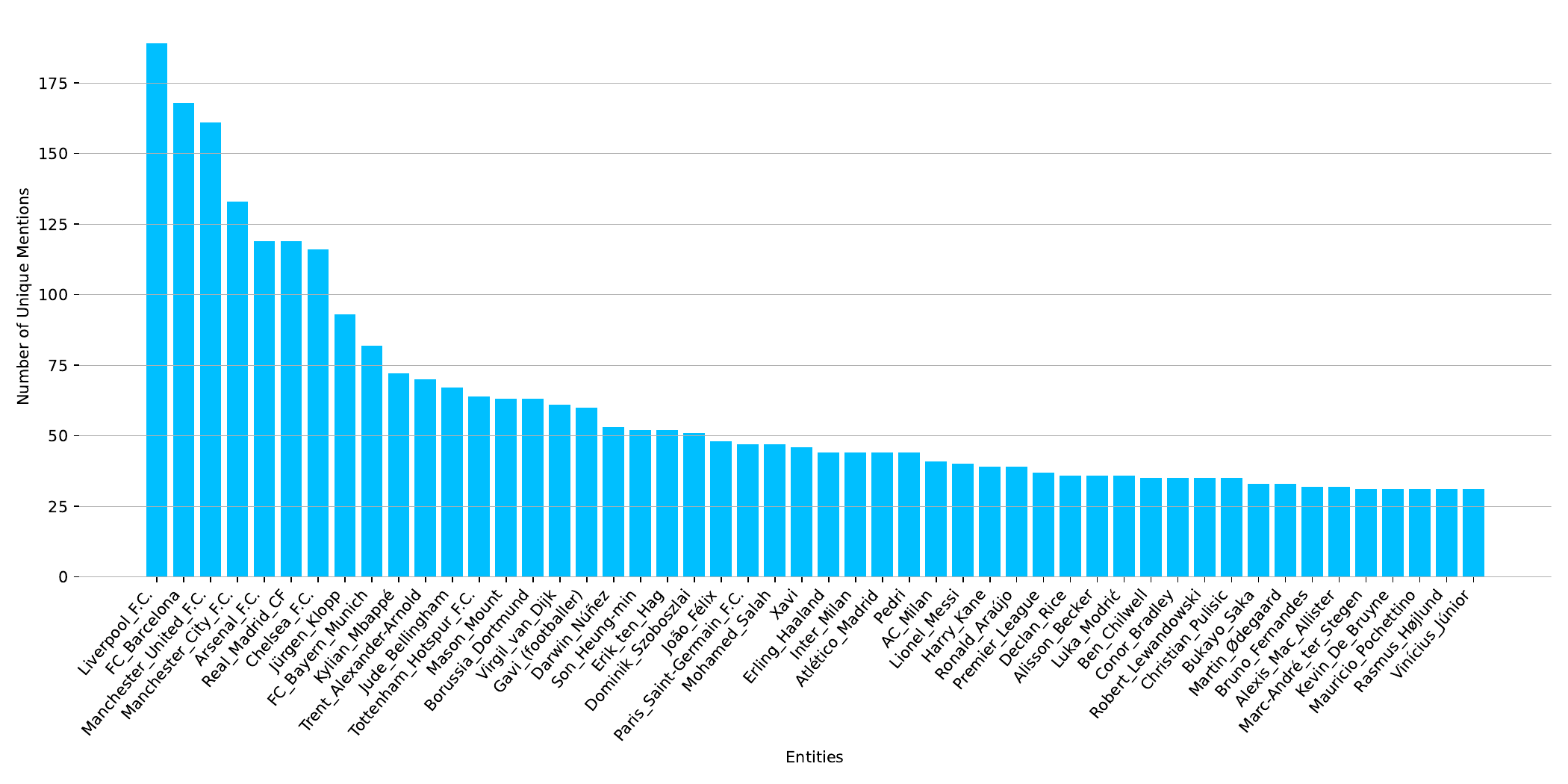}
    \caption{
    Mention variations in \textproc{DynamicER}
    }
    \label{fig:mention_variation}
\end{figure*}

\begin{table*}
\small
\renewcommand{\arraystretch}{1.4}
\resizebox{\textwidth}{!}{
\begin{tabular}{!{\vrule width 1.2pt} p{\linewidth} !{\vrule width 1.2pt}}
    \Xhline{1.2pt}
        This context involves a social media post regarding sports: Association football (Baseball).\\
        Please review the expressions that GPT-4 highlighted, which include direct names, nicknames, and role-specific references implying a specific individual or team.\\
        If any of the highlighted expressions are incorrect, please revise them.\\
        After revising, link each expression to its corresponding Wikipedia entity. You can search for these entities on Wikipedia and provide the URL. If the context is too ambiguous to resolve, or no suiable Wikipedia entity exists, label it as NOT VALID.\\
    \hline
        The provided mention \textit{\{mention\}} refers to \textit{\{entity\}} in the text.\\
        Please determine whether the mention unambiguously refer to \textit{\{entity\}} without context.\\
    \Xhline{1.2pt}
\end{tabular}
}
\caption{Full text of instructions}
\label{tab:fulltext1}
\end{table*}
\newpage
\section{Case Study}

Tables~\ref{tab:case_study_qa1},~\ref{tab:case_study_qa2}, and~\ref{tab:case_study_qa3} present a case study from entity-centric QA.

\begin{table*}
\small
\renewcommand{\arraystretch}{1.4}
\resizebox{\textwidth}{!}{
\begin{tabular}{p{\linewidth}}
    \Xhline{1.0pt}
        Question: When did \textit{Targett} make his first Premier League start? \\
        Gold Answer: December 28, 2014. \\ 
        Gold Entity: Matt Targett \\
        \hline
        LLM: \textbf{1992}. (F1 score: 0)\\
        LLM-ER: Targett made his first Premier League start for Southampton against Manchester United on \textbf{August 16, 2014}. (F1 score: 0.11 / Entity Linking: Succeed)\\
        RaLM: Targett made his first Premier League start for Aston Villa on \textbf{23 September 2014}. (F1 score: 0.12 / Retrieval: Hit)\\
        RaLM-CoT: Targett made his first Premier League start for Aston Villa on \textbf{23 September 2014}, in a 2-1 victory at ... (F1 score: 0.09 / Retrieval: Hit) \\
        RaLM-ER: Targett made his first Premier League start for Southampton on \textbf{28 December 2014}, in a 1–1 draw against Chelsea. (F1 score: 0.29 / Retrieval: Hit / Entity Linking: Succeed) \\
    \Xhline{1.0pt}
\end{tabular}
}
\caption{Case: Properly retrieved document with correctly resolved entity synergies in RaLM-ER.}
\label{tab:case_study_qa1}
\end{table*}

\begin{table*}
\small
\renewcommand{\arraystretch}{1.4}
\resizebox{\textwidth}{!}{
\begin{tabular}{p{\linewidth}}
    \Xhline{1.0pt}
        Question: When did \textit{Caoimhin the Prettiest of all Kelleher} sign his long-term deal with the club? \\
        Gold Answer: Caoimhín Kelleher signed his long-term deal on 24 June 2021. \\
        Gold Entity: Caoimhín Kelleher \\
        \hline
        LLM: Caoimhin Kelleher has not signed a long-term deal with the club. (F1 score: 0.42)\\
        LLM-ER: Caoimhin the Prettiest of all Kelleher signed his long-term deal with the club in 2021. (F1 score: 0.52 / Entity Linking: Succeed)\\
        RaLM:  In August 2018. (F1 score: 0 / Retrieval: Miss) \\
        RaLM-CoT: In August 2018, Caoimhín Odhrán Kelleher signed a new contract with Liverpool. (F1 score: 0.29 / Retrieval: Miss)\\
        RaLM-ER:  Caoimhin the Prettiest of all Kelleher signed his long-term deal with the club in 2021. (F1 score: 0.52 / Retrieval: Miss / Entity Linking: Succeed) \\
    \Xhline{1.0pt}
\end{tabular}
}
\caption{Case: Correctly resolved entity refrains from hallucinations in RaLM-ER.}
\label{tab:case_study_qa2}
\end{table*}

\begin{table*}
\small
\renewcommand{\arraystretch}{1.4}
\resizebox{\textwidth}{!}{
\begin{tabular}{p{\linewidth}}
    \Xhline{1.0pt}
        Question: What record did \textit{Trentnation} set after Liverpool's victory in the 2022 FA Cup Final? \\
        Gold Answer: He became the youngest-ever player to win six major trophies for an English club at the age of 23 years and 219 days. \\
        Gold Entity: Trent Alexander-Arnold \\
        \hline
        LLM: Trent Alexander-Arnold set a record for being the youngest player to win the FA Cup, FA Cup Final Man of the Match, and FA Cup (F1 score: 0.29)\\
        LLM-ER: Trentnation set the record for the most assists in a single season in the Premier League with 13 assists. (F1 score: 0.06 / Entity Linking: Succeed)\\
        RaLM: None, as Trent Alexander-Arnold did not set any record in the 2022 FA Cup Final. (F1 score: 0 / Retrieval: Miss) \\
        RaLM-CoT: There is no record set by Trentnation (Trent Alexander-Arnold) in the 2022 FA Cup Final. (F1 score: 0 / Retrieval: Miss)\\
        RaLM-ER:  Trent Alexander-Arnold became the youngest player to lift the FA Cup, at the age of 24. (F1 score: 0.36 / Retrieval: Miss / Entity Linking: Succeed) \\
    \Xhline{1.0pt}
\end{tabular}
}
\caption{Case: Correctly resolved entity refrains from hallucinations in RaLM-ER.}
\label{tab:case_study_qa3}
\end{table*}

\end{document}